\documentclass[10pt,twocolumn,letterpaper]{article}

\usepackage[pagenumbers]{cvpr} 

\usepackage[dvipsnames]{xcolor}

\definecolor{cvprblue}{rgb}{0.21,0.49,0.74}
\usepackage[pagebackref,breaklinks,colorlinks,citecolor=cvprblue]{hyperref}

\def\paperID{*****} 
\def\confName{CVPR}
\def\confYear{2024}

\title{Compound Expression Recognition via Multi Model Ensemble}

\author{
Jun Yu$^1$, Jichao Zhu $^1$\thanks{Corresponding author} , Wangyuan Zhu$^1$\\
$^1$University of Science and Technology of China\\
\tt\small harryjun@ustc.edu.cn\\
\tt\small \{jichaozhu,zhuwangyuan\}@mail.ustc.edu.cn \\
}

\begin{document}
\maketitle
\begin{abstract}

Compound Expression Recognition (CER) plays a crucial role in interpersonal interactions. Due to the existence of Compound Expressions , human emotional expressions are complex, requiring consideration of both local and global facial expressions to make judgments.
In this paper, to address this issue, we propose a solution based on ensemble learning methods for Compound Expression Recognition. Specifically, our task is classification, where we train three expression classification models based on convolutional networks, Vision Transformers, and multi-scale local attention networks. Then, through model ensemble using late fusion, we merge the outputs of multiple models to predict the final result. Our method achieves high accuracy on RAF-DB and is able to recognize expressions through zero-shot on certain portions of C-EXPR-DB.

\end{abstract}    
\section{Introduction}
\label{sec:intro}


In recent years, with advancements in artificial intelligence and human-computer interaction technology, automatic facial expression analysis has become a crucial research tool in fields such as clinical psychology, psychiatry, and cognitive science. It has demonstrated promising results on specific test databases and holds significant commercial prospects in various domains, including human-computer interaction\cite{Ishiguro2001RobovieAI}, virtual reality \cite{Zheng2021EyeFV} , augmented reality\cite{Bhardwaj2023EmotionDF}, smart driving\cite{Kong2015ASO}, Depression recognition\cite{Zheng2023TwoBW}. Companies like Affectiva and Kairos provide real-time assessment and prediction services, such as intelligent advertising and safe driving, by analyzing facial expressions along with other human behaviors such as language, gaze, body movements, and responses in human-computer interaction. Facial expressions have significant research value; however, in daily human life, facial expressions are not always singular in nature. They are composed of various basic expressions. For example, surprise may include both happiness and astonishment. Due to the ambiguity and diversity of expressions, researchers have started paying more attention to the recognition of compound expressions. These compound expressions are more diverse and accurately reflect the complexity and subtlety of our daily emotional expressions.

To facilitate the advancement of compound expression recognition in real-world settings, previous \cite{kollias2022eccv, kollias2022cvpr, kollias2021distribution, kollias2021analysing, kollias2021affect, kollias2020analysing, kollias2019expression, kollias2019face, kollias2019deep} and the 6th Affective Behavior Analysis in-the-wild (ABAW) \cite{kollias20246th} is hosting a challenge aimed at designing a model capable of predicting seven compound expressions. Although the competition organizers have not provided a training dataset or baseline, given that zero-shot capabilities typically lag behind supervised methods, this paper trains models using supervised methods on multiple facial datasets. At this juncture, it constitutes a classification task, with our objective being to develop a model capable of accurately predicting compound expressions.

The current classification task models can generally be categorized into two types: convolutional neural networks (CNNs) \cite{Krizhevsky2012ImageNetCW} and Transformer\cite{Vaswani2017AttentionIA} structures. ResNet \cite{He2015DeepRL} is one of the commonly used backbones in CNNs, which calculates advanced features of images by sliding convolutional kernels over them, thus focusing on local features. Vision Transformer\cite{Dosovitskiy2020AnII}, on the other hand, is the first backbone in the Transformer family to be widely applied to image modalities and achieve good results. It segments images into patches and then flattens them into sequences. By incorporating position encoding, Vision Transformer embeds the positional information of each patch into the sequence. Through Transformer's Encoder module, Vision Transformer can model the positional relationships of all positions in the image simultaneously, capturing contextual information from different parts of the face and obtaining global information. Both of these models have shown promising performance in facial expression recognition tasks in real-world scenarios.

In this paper, we employ three different models to address the issue of compound expression recognition. The convolutional network used is ResNet50, and the Transformer used is Vision Transformer, each focusing on local and global features, respectively. Additionally, we introduce the Multi-scale and Local Attention Network\cite{Zhao2021LearningDG}, which combines the two backbone concepts by incorporating multi-scale and attention mechanisms on the feature maps generated by the convolutional network.

\section{Related Work}

Facial Expression Recognition (FER) has evolved from a single expression to a composite expression, \cite{Li2023CompoundER} constructing a Real World Emotional Face (RAF-CE) database with composite expressions. Meta based multi task learning (MML) combined with AU recognition improves the performance of composite FER. Empirical research has validated the effectiveness of this method, enhancing understanding of subtle differences in complex emotions. \cite{Kollias2023MultiLabelCE} built C-EXPR-DB, a real-world dataset consisting of 400 videos annotated with 13 compound expressions. And C-EXPR-NET was proposed on this dataset, which is a multitask learning method based on facial information and AU information. By updating the model through cross entropy and KL divergence, it improves the performance of CER and AU detection (AU-D). The experimental results validated the effectiveness of C-EXPR-NET and demonstrated its generalization ability in new emotion recognition environments. In \cite {Shang2018BlendedEI}, they proposed a manifold based deep learning network called Deep Bi Manifold CNN (DBM-CNN) , which preserves the local correlation of deep features and the manifold structure of expression labels to learn the feature description of composite expressions.
\section{Method}

\begin{figure*}[htbp]
    \centering
    \includegraphics[width=\textwidth]{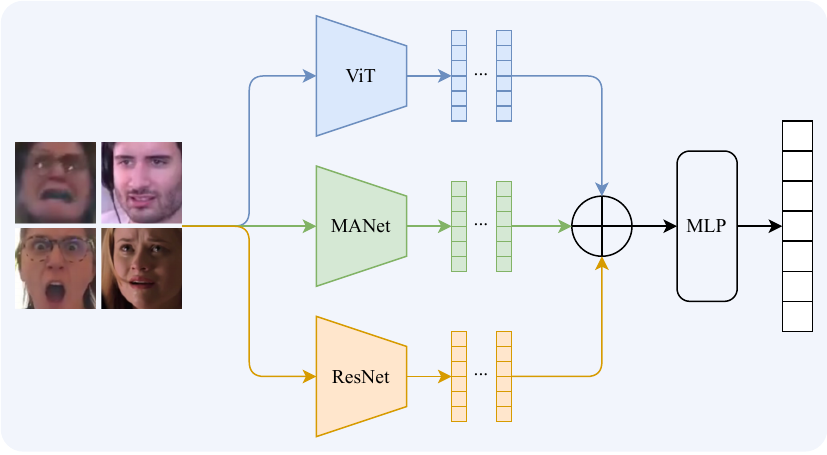}
    \caption{Ensemble}
    \label{fig:ce5}
\end{figure*}

This section we will present our late-fusion ensemble model designed to tackle compound expression recognition. The schematic of the pipeline is depicted in \cref{fig:ce5}, comprising three distinct models: Vision Transformer (ViT), MANet, and ResNet.


\subsection{Data collection}
Due to disparities between the ImageNet dataset and facial expression recognition datasets, and the limited availability of datasets with annotations for compound expressions like RAF-DB, we construct a \textbf{Unity} based on single-expression annotations from AffectNet \cite{Mollahosseini2017AffectNetAD} and RAF-DB \cite{Li2017ReliableCA}, a total of 306,989 facial images, with 299922 for training, and 7067 for validating.

\subsection{Encoders}
\subsubsection{Vision Transformer (ViT)}
We employed a pre-trained ViT from \cite{MaskedAutoencoders2021}, which underwent pre-training on the ImageNet-1K dataset using self-supervised learning with masked reconstruction. Subsequent experiments in the paper have validated its efficacy across various image-based tasks, achieving comparable or superior generalization performance compared to supervised training, including emotion recognition for facial expressions. The model processes extracted facial images and yields 768-dimensional embeddings for each image.

\subsubsection{Multi-scale and ocal Attention Network (MANet)}
The feature extractor of MANet is used for pre-extracting intermediate-level features. The multi-scale module is employed for integrating features from different receptive fields. The local attention module can guide the network to focus on locally salient features. For each frame image, both the multi-scale module and the local attention module generate final features, which are weighted according to parameters. The initialization parameters of MANet come from pre training weights on RAFDB. Both of the outputs of multi-scale module and local attention module are 512-dimensional.

\subsubsection{Residual Neural Network (ResNet)}
ResNet50 is a deep CNN and a member of the ResNet (Residual Network) series. ResNet50 consists of 50 layers of depth and adopts the idea of residual learning. It has achieved good results in tasks such as image classification and object detection, and has become one of the classic backbones. We use the weights trained on the FER2013 as initialization parameters, generating a 2048 dimensional vector for each image.

\subsection{Ensemble}

We take batched data images $x$ as input to the model, where $x \in \mathbf{R}^{B\times 3 \times H \times W}$, with $B$ denoting the batch size, $3$ representing the RGB channels, and $H$ and $W$ being the height and width of the images, respectively. Thus, the features after data augmentation can be represented as:
\begin{equation}
    \begin{aligned}
        \text{feature}_{1} &= \text{ViT}(x) \in \mathbf{R}^{B \times 768}\\
        \text{feature}_{2} &= \text{MANet}(x) \in \mathbf{R}^{B \times 1024}\\
        \text{feature}_{3} &= \text{ResNet}(x) \in \mathbf{R}^{B \times 512} \\
    \end{aligned}
\end{equation}

To enhance model performance by integrating multiple feature maps for richer feature representation, we adopt a late fusion strategy. Specifically, we concatenate the three aforementioned features along a specified dimension,  and then input the multiple feature maps into a multi-layer perceptron (MLP) and compute logit for seven compound expression using softmax, as follows:
\begin{equation}
    \begin{aligned}
    \text{feature} &= [\text{feature}_{1};\text{feature}_{2}; \text{feature}_{3}]\\
    \text{logit} &= \text{sfotmax}(\text{MLP}(\text{feature}))
    \end{aligned}
\end{equation}
where [ ; ; ] denotes the concatenation operation.


\section{Experiments}

\subsection{Dataset}

C-EXPR-DB\cite{Kollias2023MultiLabelCE} stands as the largest and most diverse in-the-wild audiovisual database to date. It encompasses 400 videos, totaling around 200,000 frames, meticulously annotated for 12 compound expressions and various affective states. Additionally, C-EXPR-DB includes annotations for continuous valence-arousal dimensions [-1, 1], speech detection, facial landmarks and bounding boxes, 17 action units (facial muscle activation), and facial attributes. In the Compound Expression (CE) Recognition Challenge, a total of 56 unlabeled videos were selected, covering 7 types of compound expressions. The extracted video tags consist of seven compound expression: Fearfully Surprised, Happily Surprised, Sadly Surprised, Disgustedly Surprised, Angrily Surprised, Sadly Fearful, and Sadly Angry.

\subsection{Implement Details}

\subsubsection{Evaluation metric}
For Compound Expression Recognition Challenge, the evaluation metric is the F1 Score for seven compound expressions, which is used to measure the prediction accuracy of the model for each expression category, combining both precision and recall, providing a more comprehensive assessment of the model's overall performance, then the metric can be defined as follows:
\begin{equation}
    F_{1} = \sum_{i=1}^{7} \frac{F_{1}^{i}}{7}
\end{equation}
where $F_1^{i}$ corresponds to the $i$-th expression.

\subsubsection{Training setting}

All experiments in this paper are conducted using the PyTorch and trained on an RTX3090. The input image resolution is consistently set to 224 $\times$ 224. During training, the number of epochs is set to 100. Cross-entropy loss is utilized as the optimization objective, and the Adam optimizer is employed for parameter updates.
In pre-train ViT on Unity, in order to improve stability during training optimization, warm-up learning rate is employed to update the learning rate.
In the RAF-DB compound expression experiment, the learning rate is set to $5e-5$, and the batch size is set to 128.

\subsection{Results}

\paragraph{Visual Models' performance on Unity}

The \cref{tab:modelunity} presents a comparison of the performance of ViT and ResNet on facial expression recognition tasks (anger, contempt, disgust, fear, happiness, neutral, sadness, and surprise), along with reporting overall accuracy (acc) and F1 scores. ViT outperforms or is comparable to ResNet in recognizing expressions of happiness, neutrality, and sadness. However, ResNet achieves slightly higher accuracy in recognizing the surprise expression. For other expressions, the performance difference between the two models is not significant. Notably, ResNet outperforms ViT by 11.02\% in accuracy on the contempt expression, indicating a clear advantage for ResNet. This may suggest that ResNet is more sensitive to capturing specific details or certain local features, which may be prominent in the contempt expression, such as eye movements or micro-expressions.

\begin{table}[htbp]
  \centering
  \caption{Visual Models' performance on Unity}
    \begin{tabular}{ccc}
    \toprule
    Single expression & ViT   & ResNet \\
    \midrule
    Anger & 66.62 & 66.31 \\
    Contempt & 11.02 & 17.84 \\
     Disgust & 45    & 46.52 \\
    Fear  & 49.48 & 54.53 \\
    Happiness & 95.91 & 93.18 \\
    Neutral & 85.85 & 76.53 \\
    Sadness & 74.95 & 70.86 \\
    Surprise & 62.97 & 65.98 \\
    \midrule
    acc   & 70.2  & 68.78 \\
    \midrule
    F1    & 64.48 & 62.37 \\
    \bottomrule
    \end{tabular}%
    
  \label{tab:modelunity}%
\end{table}%

\begin{table}[htbp]
  \centering
  \caption{Results on RAF-DB (CE) validation set.}
    \begin{tabular}{cccc}
    \toprule
    Compound Expression & ViT   & MANet & ResNet \\
    \midrule
    Angrily Surprised & 60.53 & 52.63 & 55.26 \\
    Disgustedly Surprised & 51.43 & 37.14 & 60 \\
    Fearfully Surprised & 80.17 & 81.03 & 75.86 \\
    Happily Surprised & 92.59 & 89.63 & 85.93 \\
    Sadly Angry & 84.85 & 75.75 & 84.85 \\
    Sadly Fearful & 72.72 & 63.64 & 63.64 \\
    Sadly Surprised & 38.89 & 38.89 & 55.56 \\
    \midrule
    acc   & 78.09 & 74.06 & 75.06 \\
    \midrule
    F1    & 70.25 & 63.57 & 68.19 \\
    \bottomrule
    \end{tabular}%
    
  \label{tab:rafdbce}%
\end{table}%

\paragraph{Visual Models' performance on RAF-DB (CE)}
The \cref{tab:rafdbce} illustrates the performance comparison of  ViT, MANet and ResNet in recognizing compound expressions on RAF-DB validation set. For each compound expression, the table lists the recognition accuracy of the three models, as well as their accuracy and F1 score.
In recognizing the Happily Surprised expression, ViT leads with an accuracy of 92.59\%. In the recognition of the Sadly Surprised expression, ResNet significantly outperforms ViT and MANet with an accuracy of 55.56\%, while the latter two have accuracies of 38.89\%.

In terms of overall performance, ViT leads with an accuracy of 78.09\%, followed by ResNet with 75.06\%, and MANet with 74.06\%. In terms of F1 score, ResNet and ViT perform similarly at 68.19\% and 70.25\%, respectively, while MANet has the lowest performance at 63.57\%.
\begin{figure}
    \centering
    \includegraphics[width=0.5\textwidth]{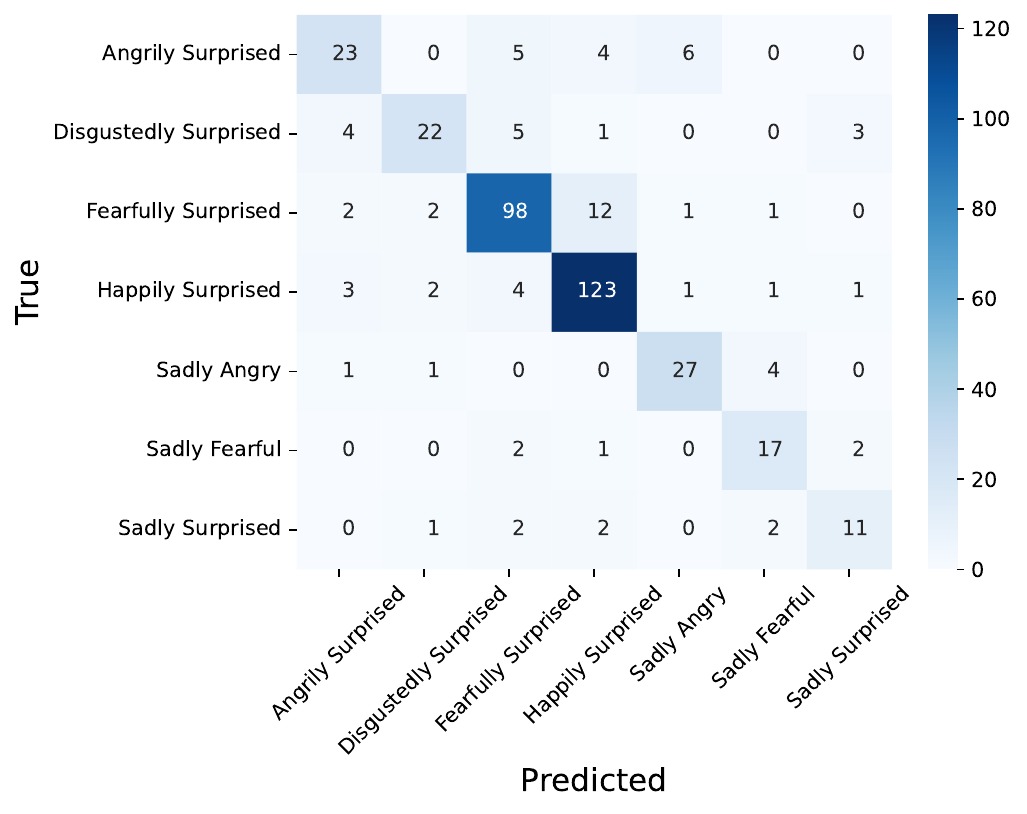}
    \caption{Confusion Matrix of Ensemble Models in RAF-DB Compound Expressions}
    \label{fig:cm}
\end{figure}

Considering both accuracy and F1 score, ViT demonstrates good balanced performance overall, although it may not be the best in some individual expressions. These results indicate that different network architectures have different strengths and weaknesses in recognizing complex emotional expressions, and no single model is optimal in all cases. This also suggests the need to combine multiple models to improve the accuracy and robustness of expression recognition in practical applications.

\begin{table}
  \centering
  \caption{Results of Ensemble Models in RAF-DB Compound Expressions}
    \begin{tabular}{cc}
    \toprule
    Compound Expression & Ensemble \\
    \midrule
    Angrily Surprised & 60.53 \\
    Disgustedly Surprised &  62.86\\
    Fearfully Surprised &  87.48\\
    Happily Surprised &  91.11\\
    Sadly Angry &  81.82\\
    Sadly Fearful &  77.27\\
    Sadly Surprised &  61.11\\
    \midrule
    acc   &  80.86\\
    \midrule
    F1    & 74.60 \\
    \bottomrule
    \end{tabular}%
  \label{tab:emce}%
\end{table}%

\paragraph{Ensemble models' performance on RAF-DB(CE)}

The results of the ensemble models with late fusion on the RAF-DB dataset are shown in \cref{tab:emce}, and it can be seen that
Compared with the single model, the integrated model is more accurate in predicting five compound expressions, namely, Angry Surprised, Disgustedly Surprised, Disgustedly Surprised, Sadly Fearful, and Sadly Surprised. In particular, the expression of Sadly Surprised, which is difficult to identify, is 22.22\% higher than that of ViT. In addition, the accuracy rate and F1 score are both better, which is consistent with our idea of using different models to bridge the gap between each other. \cref{fig:cm} illustrates the associated confusion matrix, where the diagonal cells signify the count of accurately predicted samples for each expression.

\section{Conclusion}

{
    \small
    \bibliographystyle{ieeenat_fullname}
    \bibliography{main}
}


\end{document}